\documentclass[a4paper,twoside]{article}

\usepackage{epsfig}
\usepackage{subcaption}
\usepackage{calc}
\usepackage{amssymb}
\usepackage{amstext}
\usepackage{amsmath}
\usepackage{amsthm}
\usepackage{multicol}
\usepackage{pslatex}
\usepackage{apalike}
\usepackage{array}
\usepackage{siunitx}
\usepackage{hyperref}
\usepackage{SCITEPRESS}     % Please add other packages that you may need BEFORE the SCITEPRESS.sty package.

\begin{document}

\title{Automatic Estimation of Anthropometric Human Body Measurements}

\author{\authorname{paper 189}}

\author{\authorname{Dana Škorvánková\sup{1}%\orcidAuthor{0000-0003-3791-495X}
, Adam Riečický\sup{2}%\orcidAuthor{0000-0002-1546-0048}
and Martin Madaras\sup{1,2}%\orcidAuthor{0000-0003-3917-4510}
} 
\affiliation{\sup{1}Faculty of Mathematics, Physics and Informatics, Comenius University Bratislava, Slovakia}
\affiliation{\sup{2}Skeletex Research, Slovakia}
\email{dana.skorvankova@fmph.uniba.sk, \{madaras, riecicky\}@skeletex.xyz} 
}

\keywords{Computer Vision, Neural Networks, Body Measurements, Human Body Analysis, Anthropometry, Point Clouds}

\abstract{
%The abstract should summarize the contents of the paper and should contain at least 70 and at most 200 words. The text must be set to 9-point font size.
Research tasks related to human body analysis have been drawing a lot of attention in computer vision area over the last few decades, considering its potential benefits on our day-to-day life. Anthropometry is a field defining physical measures of a human body size, form, and functional capacities. Specifically, the accurate estimation of anthropometric body measurements from visual human body data is one of the challenging problems, where the solution would ease many different areas of applications, including ergonomics, garment manufacturing, etc. This paper formulates a research in the field of deep learning and neural networks, to tackle the challenge of body measurements estimation from various types of visual input data (such as 2D images or 3D point clouds). Also, we deal with the lack of real human data annotated with ground truth body measurements required for training and evaluation, by generating a synthetic dataset of various human body shapes and performing a skeleton-driven annotation.} 

\onecolumn \maketitle \normalsize \setcounter{footnote}{0} \vfill

\section{\uppercase{Introduction}}
\label{sec:introduction}

Analyzing human body and motion has been an important field of research for decades. The related tasks attract attention of many computer vision researchers, mainly due to the wide range of applications, which includes surveillance, entertainment industry, sports performance analysis, ergonomics, human-computer interaction, garment manufacturing, etc.

Human body analysis covers a number of different tasks, including body parts segmentation, pose estimation and body measurements estimation; while capturing human body in motion brings up additional tasks, such as pose tracking, activity recognition and classification, and many more. All of the topics are closely related, thus are often treated as associated or complementary tasks.

Anthropometric human body measurements gather various statistical data about human body and its physical properties. They are generally categorized into two groups: static and dynamic measurements. Static, or structural, dimensions include circumferences, lengths, skinfolds and volumetric measurements. Dynamic, or functional, dimensions incorporate link measurements, center of gravity measurements, and body landmark locations. In this research, we will focus mainly on body circumferences, widths and lengths of particular body parts or limbs, and other distances within a human body. One of the issues in context of anthropometry is the lack of standardization in body measurements. For this reason, we clarify the definition of each annotated measurement in Section~\ref{subsec:data_anot}.

For the purpose of the anthropometric body measurements estimation, there has been very few data with ground truth annotations made publicly available. Up to our knowledge, the only large-scale dataset of real human body scans along with the manually measured body dimensions is a commercial dataset CAESAR~\cite{Robinette2002}, which has not been released for public usage. Since annotating real data using tape measuring is rather exhausting and time-consuming, considering the potentially large set of different human subjects; the usual workaround is to make use of synthetically generated data. However, at the cost of the relatively fast data annotation, there is a need for establishing a robust method to obtain the accurate body measures on the body surface.

The main contribution of this paper is fourfold: (1) we examine various 2D and 3D input human body data representations, along with their impact on the ability of a neural network to extract the important features and process them to estimate true values of a predefined set of body measurements on the output; (2) to deal with the insufficient amount of publicly available data annotated with ground-truth body measurements, we generated a large-scale synthetic dataset of various body shapes in standard body pose, using parametric human body model, along with corresponding point clouds, gray-scale and silhouette images, skeleton data, and 16 annotated body measurements; (3) to obtain the ground-truth for the 16 measurements on the body models, we established a skeleton-guided annotation pipeline, which can easily be extended to compute more complex and task-specific body dimensions, and finally, (4) we present a method for an accurate automatic end-to-end human body measurements estimation from a single input frame.

\begin{figure*}
\centering
\includegraphics[width=\textwidth]{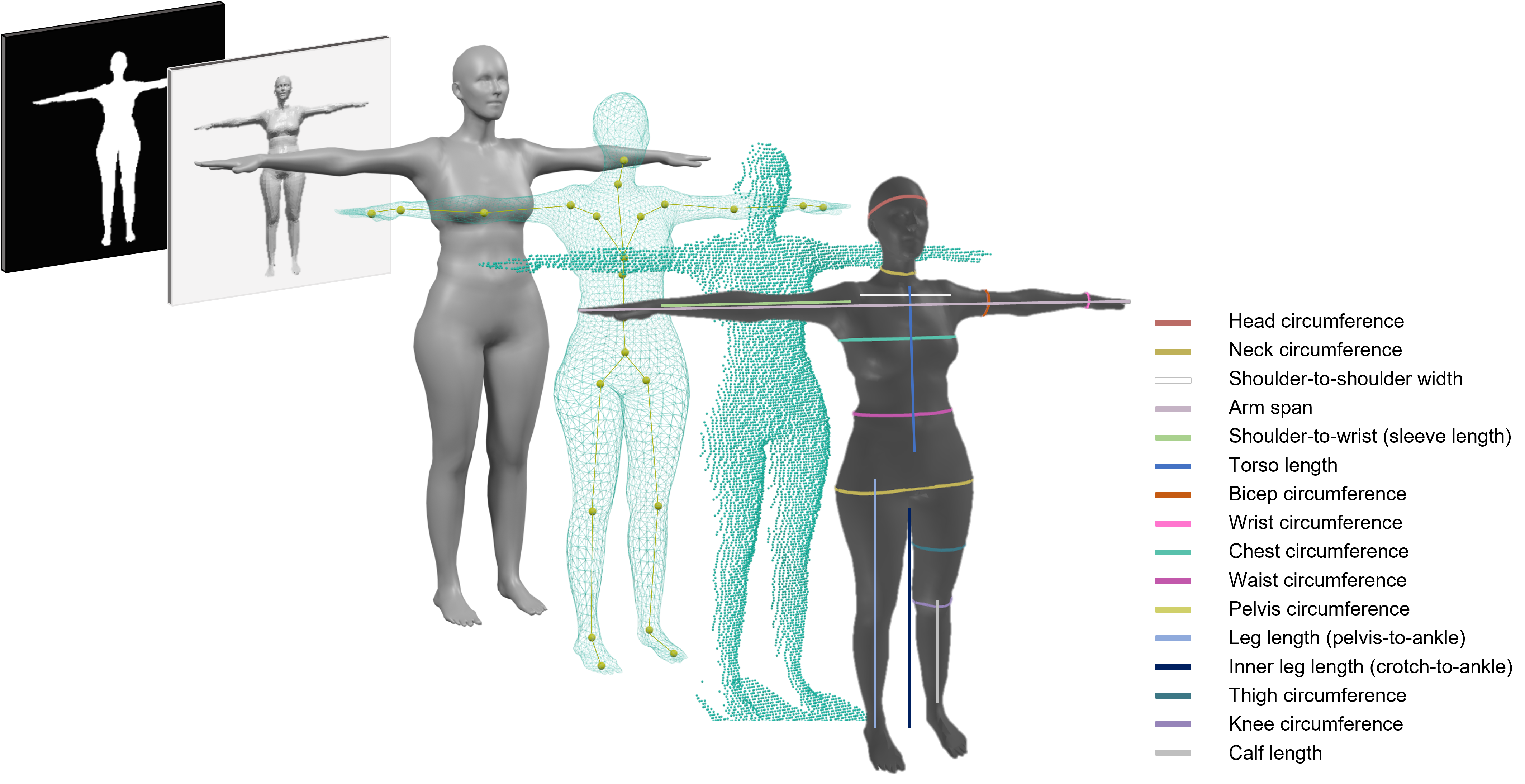}
\caption{Human data representations included in the generated synthetic dataset: binary silhouette images, gray-scale images, 3D models, skeleton data, 3D point clouds and 16 annotated body measurements.}
\label{fig:dataset}
\end{figure*}

\section{\uppercase{Related work}}

The anthropometric body measurements estimation is an emerging problem in the context of various applications, such as garment manufacturing, ergonomics, or surveillance. An automatic estimation of accurate body measures would prevent us from having to manually tape measure the human bodies. Also, the automated pipeline would bring consistency in body measuring, which is often hard to maintain when tape measuring different human subjects. Aside from natural human error, or inaccuracies caused by tape measuring, there is an ambiguity across various different body measuring standards.

There have been numerous algorithmic strategies presented to tackle the task of human body measurements estimation over the years~\cite{Guillo2020,Anisuzzaman19,ASHMAWI2019,Song2017,Dao2014,Tsoli2014,Li2013}. However, they often proved not to satisfy the accuracy of the estimations, nor meet the desired efficiency, or computational and complexity requirements. One of the main problems when processing data representing human body is the irregularity and complex structure of the human body surface. In theory, there are no predefined vertices on the surface of human body to guide the processing; as it is when analyzing standard 3D objects with corners and edges. Considering the theoretical and practical issues with the algorithmic approaches, in many environments, they have been replaced with machine learning techniques, such as random forests~\cite{XIAOHUI2018} or neural networks~\cite{Yan2021,Wang2019}. 

In order to sufficiently train a machine learning model, a large amount of human body data annotated with ground truth body measurements is essential. In general, there are no such large-scale benchmark datasets publicly available for research purposes at the moment. The main reason for this is the exhausting process of manual tape measuring of real human bodies. Therefore, most researchers have made use of the synthetic data instead of the real human data. Tejeda et al.~\cite{Tejeda2019} focused on the annotation process of three basic body measurements: chest, waist, and pelvis circumference on 3D human body model. Our annotation method presented in this paper is inspired by their approach, while we optimized and adjusted the conditions in computation of the particular measurements, and extended the set of measurements by thirteen additional body measures.

\subsection{1D Statistical Input Data} 
Regarding the human body measurements estimation, several existing approaches formulate the task as estimating an extended list of advanced body measures from a set of predefined basic body measurements on the input, thus having the 1D statistical input data~\cite{Wang2019,Liu2017}. Usually, the estimation is based on an end-to-end learning neural network, mapping from the input \textit{easy-to-measure} body dimensions to the detailed body dimensions on the output. However, these methods still require manual tape measuring of the few basic attributes, which may be inconvenient in certain application scenarios.

\subsection{Image Input Data}
Methods inferring from 2D input images were proposed to estimate the body measurements from visual data to avoid the need for manual measuring in deployment. Most frequently, the input data are in a form of RGB images~\cite{Yan2021,Anisuzzaman19,Shigeki2018}, although the three color channels may not be very beneficial in context of this particular task, at the cost of processing the three-channeled data. Thus, several other approaches settled for gray-scale images~\cite{Tejeda2021} as input data, while achieving competitive results. 

Binary silhouette images of a human body were also used in some of the strategies~\cite{Tejeda2019,Song2017}, suggesting the contours of the body shape are the most important feature for the stated task.

\subsection{3D Input Data}
Furthermore, there has been a number of methods proposing the engagement of 3D input data~\cite{Guillo2020,XIAOHUI2018,Tsoli2014}. In~\cite{Yan2020}, the authors are fitting a Skinned Multi-Person Linear (SMPL) body model to a scanned point cloud, using a non-rigid iterative closest point algorithm as a part of their pipeline. Then, they run a non-linear regressor to estimate the body measures on the fitted body model from multiple measured circumference paths. There also have been few attempts to compute the body dimensions on 3D body point clouds~\cite{Dao2014} using analytical approaches, although the idea has not been developed much further.

One of the main contributions of this paper are experiments with 3D input data, where we suggest using 3D point clouds directly on the input, and training a neural model in an end-to-end fashion to avoid the need for an expensive alignment of the point cloud and 3D body model.

\section{\uppercase{Proposed Approach}} 
In this section, we present our proposed strategy to accurately estimate the anthropometric body measurements from visual data. In our work, we examine various input data types, including binary silhouette images, gray-scale images and 3D point clouds. Another contribution of our research is an established framework for a skeleton-guided computation of 16 ground-truth body measurements on a 3D body model. We have produced a large-scale database of synthetic human body data relevant for various human body analysis-related tasks and we publish the dataset\footnote{\url{http://skeletex.xyz/portfolio/datasets}} for further research.

\subsection{Data Acquisition}
We consider generating a large-scale collection of synthetic human body data one of the contributions of this paper, while containing multiple corresponding data representations, categorized into male and female body data. It includes 3D body models, point clouds, gray-scale and binary silhouette images, skeleton data as well as a set of 16 annotated body measurements, as illustrated in Figure~\ref{fig:dataset}. We present the details of the database in the subsequent sections.

\setlength{\tabcolsep}{4pt}
\begin{table*}[t]
\caption{Definition of annotated anthropometric body measurements. Note that the 3D model is expected to capture the human body in the default T-pose, with Y-axis representing the vertical axis, and Z-axis pointing towards the camera.}
\label{table:measurements}
\small
\begin{tabular}{p{0.2\textwidth}p{0.8\textwidth}}
\hline\noalign{\smallskip}
Body measurement & Definition \\
\noalign{\smallskip}
\hline
\noalign{\smallskip}
Head circumference & circumference taken on the Y-axis at the level in the middle between the head skeleton joint and the top of the head (the intersection plane is slightly rotated along X-axis to match the natural head posture)\\
Neck circumference & circumference taken at the Y-axis level in 1/3 distance between the neck joint and the head joint (the intersection plane is slightly rotated along X-axis to match the natural posture)\\
Shoulder-to-shoulder & distance between left and right shoulder skeleton joint\\
Arm span & distance between the left and right fingertip in T-pose (the X-axis range of the model)\\
Shoulder-to-wrist &  distance between the shoulder and the wrist joint (sleeve length)\\
Torso length & distance between the neck and the pelvis joint\\
Bicep circumference & circumference taken using an intersection plane which normal is perpendicular to X-axis, at the X coordinate in the middle between the shoulder and the elbow joint\\
Wrist circumference & circumference taken using an intersection plane which normal is perpendicular to X-axis, at the X coordinate of the wrist joint\\
Chest circumference & circumference taken at the Y-axis level of the maximal intersection of a model and the mesh signature within the chest region, constrained by axilla and the chest (upper spine) joint\\
Waist circumference & circumference taken at the Y-axis level of the minimal intersection of a model and the mesh signature within the waist region -- around the natural waist line (mid-spine joint); the region is scaled relative to the model stature\\
Pelvis circumference & circumference taken at the Y-axis level of the maximal intersection of a model and the mesh signature within the pelvis region, constrained by the pelvis joint and hip joint\\
Leg length & distance between the pelvis and ankle joint\\
Inner leg length & distance between the crotch and the ankle joint (crotch height); while the Y coordinate being incremented, the crotch is detected in the first iteration after having a single intersection with the mesh signature, instead of two distinct intersections (the first intersection above legs)\\
Thigh circumference & circumference taken at the Y-axis level in the middle between the hip and the knee joint\\
Knee circumference & circumference taken at the Y coordinate of the knee joint\\
Calf length & distance between the knee joint and the ankle joint\\
\hline
\end{tabular}
\end{table*}
\setlength{\tabcolsep}{1.4pt}

\subsubsection{Synthetic Data Generation}
The synthetic human body data were generated using a SMPL parametric model~\cite{SMPL2015} with high variety of different body shapes. It includes 50k male and 50k female body models. The gray-scale and binary images were generated by capturing the rendered body model from a frontal-view. Furthermore, each rendered body model was virtually scanned from two viewpoints, thus producing a corresponding 3D body point cloud containing $88\,408$ distinct points. Also, Gaussian noise is added to the virtual scans, following~\cite{Jensen2021,Rakotosaona2019,Rosman2012}, to bring the resulting data closer to the real data captured by a structured-light 3D scanner. The body scans are originally structured in 2D grid, similar to standard image, where the grid contains 3 real world coordinates at indices containing a valid point of the point cloud, and zeros (representing an empty grid index) otherwise. 
However, to merge scans from two camera viewpoints and thus incorporate more information into a single point cloud, we discard the grid structure, and use unorganized point clouds directly as an input. Another reason to prefer the unstructured point clouds would be to save the computation time and memory, related to the large number of empty points in the grid-structured scans. Nonetheless, the structured point cloud representation might be useful in certain cases, and we plan to incorporate this data format in the future steps of this research, as we suggest in Section~\ref{subsec:futurework}.

\subsubsection{Annotation Process}\label{subsec:data_anot}
In this section, we describe the process of annotating the synthetically generated data with various body dimensions measured on the body model surface. Our annotation method is inspired by CALVIS~\cite{Tejeda2019}. While we used the same mesh signature approach, intersecting the model geometry; we extended the list of body measurements and adjusted the three original measures (chest, waist and pelvis circumference) to make them consistent with the real measures obtained by manual tape measuring. In particular, we optimized the vertical axis conditions on the body region, where each of the circumference is measured. We propose, that the vertical range of each region should be relative to the stature of the specific body model. Moreover, we mark thirteen additional measurements (as described in Figure~\ref{fig:dataset}), which are often used mainly in garment manufacturing. In Table~\ref{table:measurements}, we define each of the measurements, to avoid any ambiguities considering different anthropometric measuring standards.
\vfill

\subsection{Baseline Models}
Here, we present the baseline models used in our experiments to regress the anthropometric body measurements. Our research is focused on examining various input data types, including both 2D and 3D human body data representations. 

\begin{figure*}
\centering
\includegraphics[width=\textwidth]{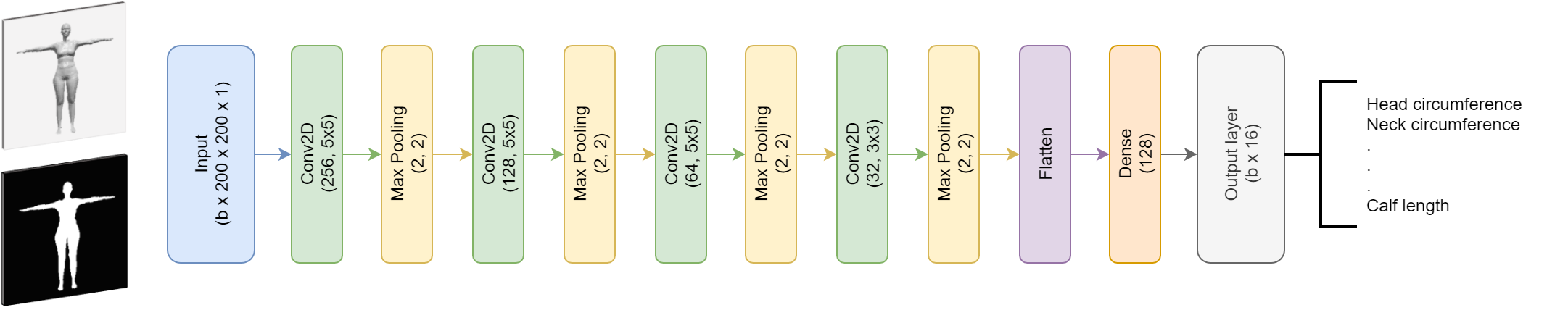}
\caption{The architecture of Convolutional Body Dimensions Estimation (Conv-BoDiEs) network. The model takes a single $200\times200$ gray-scale or binary image as input, and returns 16 estimated body measurements on the output.}
\label{fig:convnet}
\end{figure*}

\subsubsection{2D Input Data}
For 2D input data, namely gray-scale images and binary silhouette images, we employ a baseline model for Convolutional Body Dimensions Estimation (Conv-BoDiEs). The model takes a single gray-scale or silhouette image of size $200\times200$ pixels as input, same as in~\cite{Tejeda2019,Tejeda2021}; and regresses the values of 16 predefined body measurements. The network architecture is described in Figure~\ref{fig:convnet}, while all of the convolution layers are followed by ReLU activation. 

\begin{figure*}
\centering
\includegraphics[width=\textwidth]{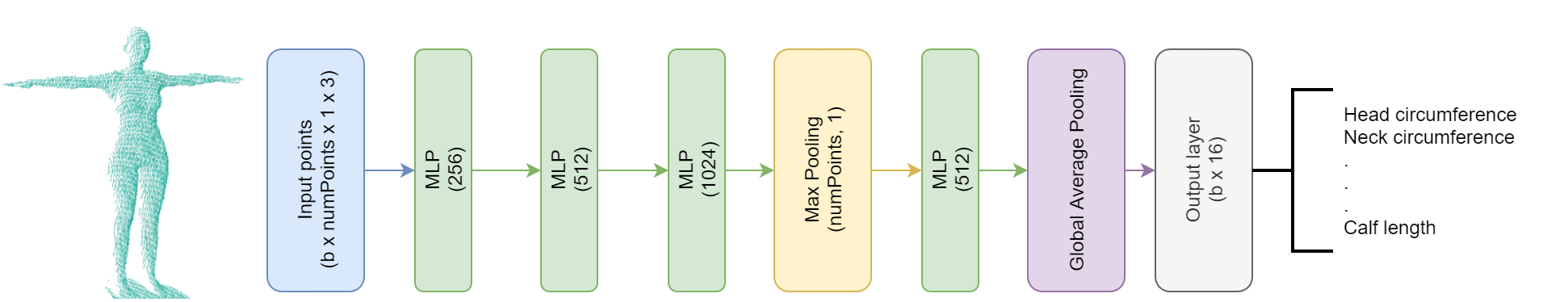}
\caption{The architecture of Point Cloud Body Dimensions Estimation (PC-BoDiEs) network. The model takes an unorganized 3D body scan merged from two viewpoints as input, and returns 16 estimated body measurements on the output. Note, that the number of points in the body scan is a hyperparameter.}
\label{fig:pclnet}
\end{figure*}

\subsubsection{3D Input Data}
In contrast to some of the previous approaches, instead of the exhausting process of fitting a body scan to a predefined body model, we aim to directly regress the body measurements from a single unorganized 3D point cloud merged from two camera viewpoints, or even a grid-structured point cloud in the future (as explained in Section~\ref{subsec:futurework}).
One of the effective methods to process the unstructured 3D body scans is to extract both global and local features in the network, and aggregate these features to maintain the information on the overall context as well as the local neighbourhoods, as formulated in~\cite{PointNet2016} and the follow-up research. Therefore, we propose a baseline neural architecture for Point Cloud Body Dimensions Estimation (PC-BoDiEs) based on stacked multi-layer perceptron (MLP) convolutions to regress the lengths of 16 stated body measurements. Details of the model architecture are shown in Figure~\ref{fig:pclnet}. Each of the MLP layers is followed by ReLU activation.

To lower the density of the body scans and thus lower the time and memory requirements of the model, the input point clouds are sub-sampled using farthest point sampling to match a fixed number of points before being fed to the network. In experiments, we validate two different point cloud densities and show the trade-off between the number of points present and the estimation accuracy of the model.

\setlength{\tabcolsep}{4pt}
\begin{table*}
\begin{center}
\caption{The quantitative results of Conv-BoDiEs and PC-BoDiEs. G means gray-scale, B means binary input image. Mean absolute error (MAE) is reported per each body measurement over all $k=5$ folds, as well as averaged over all measurements and all folds. Average Precision (AP) is displayed with two thresholds: at \SI{20}{mm} (AP@20) and at \SI{10}{mm} (AP@10). For each measurement, it illustrates the percentage of samples, where the particular measurement was estimated within the threshold from ground truth. In the last row, mean average precision shows the percentage of samples estimated with MAE under stated threshold (note, that in this case, it is not equal to the average of the above rows).} 
\label{table:results}
\small
\begin{tabular}{l|rrrr|rrrr|rrrr}
\hline
&&&&&&&&&&&&\\[-0.9\medskipamount]
Body measurement &  \multicolumn{4}{c|}{MAE (mm)} & \multicolumn{4}{c|}{AP@20 (\%)} & \multicolumn{4}{c}{AP@10 (\%)} \\
&&&&&&&&&&&&\\[-0.9\medskipamount]
 & \multicolumn{2}{c}{Conv-BoDiEs} &  \multicolumn{2}{c|}{PC-BoDiEs} &  \multicolumn{2}{c}{Conv-BoDiEs} &  \multicolumn{2}{c|}{PC-BoDiEs} &  \multicolumn{2}{c}{Conv-BoDiEs} &  \multicolumn{2}{c}{PC-BoDiEs}\\
 & G & B & & & G & B & & & G & B & & \\
% \noalign{\smallskip}
\hline
&&&&&&&&&&&&\\[-0.9\medskipamount]
% \noalign{\smallskip}
Head circumference & 8.38 &16.22 & & 8.06 & 94.09& 67.56& &94.87 &66.12 &37.70 & &68.44\\
Neck circumference & 8.82 & 17.39& & 9.07& 93.08& 64.54& & 91.76& 63.81&35.57 &&62.46\\
Shoulder-to-shoulder &7.54 &12.41 & &8.21 & 96.37& 80.36& & 94.57& 71.28&48.06&&67.43\\
Arm span & 5.32 & 7.45 & & 6.95& 99.63& 96.82& & 97.75& 86.77& 71.88&&75.57\\
Shoulder-to-wrist &3.90 & 6.00 & &5.18 &99.97 &99.14 & &99.66 &95.81 & 81.67&&87.63\\
Torso length &6.51 & 10.13& & 7.85& 98.46& 88.48& & 95.68&78.10 & 56.99&&69.23\\
Bicep circumference & 4.60& 6.66& & 5.79& 99.87& 98.37& & 99.40&91.46 &77.05 &&83.16\\
Wrist circumference &2.23 & 3.28& & 2.48&100.00 & 99.99& & 100.00& 99.80& 98.11&&99.79\\
Chest circumference  & 2.57& 5.24& & 3.29& 100.00& 99.71& &100.00 & 99.57& 87.22&&98.31\\
Waist circumference & 1.65& 3.11& &2.29 &100.00 & 100.00& &100.00 & 99.98& 98.96&&99.96\\
Pelvis circumference & 3.51 & 4.92& & 5.11&99.89 &99.57 & &99.66 &97.09 & 89.52&&88.17\\
Leg length & 2.65& 3.69& & 3.48& 100.00& 100.00& & 100.00&99.63 & 96.97&&97.77\\
Inner leg length &4.16 &5.80 & &2.76 & 99.67& 98.51& & 99.99& 94.10&83.89 &&98.92\\
Thigh circumference & 2.46&3.31  & &2.80 &99.99 &99.97 & & 99.99& 99.75& 97.98&&99.41\\
Knee circumference &2.76 &5.47 & & 3.45&99.98 & 99.47& & 99.98& 99.33& 85.38&&97.67\\
Calf length & 7.27& 10.56& &7.90 &96.08 & 87.39& &95.20 &73.23 & 53.68&&69.11\\
\hline
% \noalign{\smallskip}
&&&&&&&&&&&&\\[-0.9\medskipamount]
Mean & \textbf{4.64} & 7.60 & & 4.95 & \textbf{100.00} & 99.99 & & \textbf{100.00} & 99.84 & 88.70 & & \textbf{99.86}\\
% \noalign{\smallskip}
\hline
\end{tabular}
\end{center}
\end{table*}
\setlength{\tabcolsep}{1.4pt}

\section{\uppercase{Experiments}}
In this section, we illustrate the conducted experiments using the purposed neural models on the generated dataset.
\subsection{Training Setup}
Prior to training, the whole set of gray-scale images was normalized to zero-mean and one standard deviation.
The point clouds were globally normalized to fit the range of $[-1,1]$. While sub-sampling the resolution, we conduct experiments with two settings: using 512 and 1024 points per point cloud (as reported in Table~\ref{table:results2}). 
During the training stage, the loss function used for both stated networks was mean absolute error. All results are reported after training the models for 300 epochs. For both models, the learning rate was gradually decreased using a cosine decay, with the initial value set to \num{e-4} and \num{5e-4}, for the image and point cloud network respectively. The models were trained using AMSGrad variant of the Adam optimizer, with batches of 32 samples. The experiments were conducted on Nvidia GeForce RTX 3060.

\setlength{\tabcolsep}{4pt}
\begin{table*}
\begin{center}
\caption{Performance of the PC-BoDiEs model using various input point cloud density.} 
\label{table:results2}
\small
\begin{tabular}{l|rr|rr|rr}
\hline
&&&&&&\\[-0.9\medskipamount]
Body measurement &  \multicolumn{2}{c|}{MAE (mm)} &  \multicolumn{2}{c|}{AP@20 (\%)} &  \multicolumn{2}{c}{AP@10 (\%)}\\
& 512 pts & 1024 pts & 512 pts & 1024 pts & 512 pts & 1024 pts\\
\hline
&&&&&&\\[-0.9\medskipamount]
Head circumference & 8.06 & 7.54& 94.87&100.00 &68.44 &100.00\\
Neck circumference & 9.07&8.44 & 91.76&100.00 & 62.46&100.00\\
Shoulder-to-shoulder & 8.21&7.93 & 94.57& 100.00& 67.43&100.00\\
Arm span & 6.95&6.45 &97.75 & 99.98& 75.57&99.97\\
Shoulder-to-wrist &5.18 & 4.65& 99.66& 100.00&87.63 &100.00\\
Torso length &7.85 & 7.51& 95.68&100.00 & 69.23&100.00\\
Bicep circumference &5.79 &5.51 & 99.40& 100.00& 83.16&100.00\\
Wrist circumference & 2.48& 2.32&100.00 & 100.00&99.79 &100.00\\
Chest circumference &3.29 & 2.96& 100.00&100.00 & 98.31&100.00\\
Waist circumference &2.29 &2.16 &100.00 &100.00 & 99.96&100.00\\
Pelvis circumference & 5.11& 4.80&99.66 & 100.00&88.17 &99.97\\
Leg length &3.48 &3.23 &100.00 & 100.00&97.77 &99.99\\
Inner leg length &2.76 & 2.43& 99.99&100.00 & 98.92&100.00\\
Thigh circumference &2.80 & 2.57& 99.99&100.00 &99.41 &100.00\\
Knee circumference &3.45 & 3.15& 99.98& 100.00&97.67 &100.00\\
Calf length & 7.90& 7.48& 95.20& 100.00& 69.11&99.99\\
\hline
&&&&&&\\[-0.9\medskipamount]
Mean & 5.29 & \textbf{4.95} & \textbf{100.00} & \textbf{100.00} & 99.77& \textbf{99.86}\\
\hline
\end{tabular}
\end{center}
\end{table*}
\setlength{\tabcolsep}{1.4pt}

\subsection{Evaluation}
For evaluating the models, we use k-fold validation with $k=5$. Each time, the dataset containing a total of 100k samples was split to train and test set with ratio 80:20. 

We report three evaluation metrics: (1) mean absolute error (MAE) denotes the average error between the ground truth and the predicted measurements in millimeters; (2) average precision (AP) for each measurement marks the percentage of samples where the particular measurement was estimated within the specified threshold from ground truth; and (3) mean average precision (mAP) marks the percentage of samples estimated with MAE under the stated threshold.

In Table~\ref{table:results}, we present the performance of our models Conv-BoDiEs and PC-BoDiEs which use various input data representations to estimate the value of 16 pre-defined body measurements. As shown in the table, the lowest MAE of \SI{4.64}{mm} was achieved using gray-scale input images. The MAE obtained using unorganized point cloud input data is not far behind, with the value of \SI{4.95}{mm}. In both cases, all of the 20k test samples were estimated with MAE (averaged over all body measurements) within \SI{20}{mm} from ground truth. The biggest error among particular body measurements in all scenarios was reported on neck circumference, head circumference and shoulder-to-shoulder distance; while the models performed best on waist and wrist circumferences.

\section{\uppercase{Conclusions}}\label{sec:conclusion}
In this paper, we examined various human body data representations, including 2D images and 3D point clouds, and their impact on a neural network performance estimating anthropometric body measurements. As a part of the research, we generated large-scale synthetic dataset of multiple corresponding data formats, which is publicly available for research purposes, and can be used in many human body analysis-related tasks. We introduced an annotation process to obtain ground truth for 16 distinct body measurements on a 3D body model. Finally, we presented baseline end-to-end methods for accurate body measurements estimation from 2D and 3D body input data. The results of our experiments have shown that both the grid structure and the depth information in the input data hold an important additional value, and have positive effect on final estimation. Approaches engaging grid-structured gray-scale images, as well as unstructured 3D point clouds both yield competitive results, reaching the mean error of approximately \SI{5}{mm}.

\subsection{Future Work}\label{subsec:futurework}
In the next step of our research, we plan to incorporate structure into the 3D body scans, merging the benefits of grid-structure and the depth information, and examine its impact on the model inference. In this type of data, the 3D points are organized in a 2D grid, analogously to the image grid structure. However, in structured point clouds, instead of RGB or intensity values, the three channels for each point preserve its 3D coordinates. Besides, we aim to further extend the experiments with stated data representations, and capture relevant real human data using 3D scanners, annotated with tape measured body dimensions to evaluate the models accuracy in real-world scenarios.

% \vfill
% \section*{\uppercase{Acknowledgements}} 

\bibliographystyle{apalike}
{\small
\bibliography{example}}

\begin{thebibliography}{}

\bibitem[Anisuzzaman et~al., 2019]{Anisuzzaman19}
Anisuzzaman, D.~M., Shaiket, H. A.~W., and Saif, A. (2019).
\newblock Online trial room based on human body shape detection.
\newblock {\em International Journal of Image, Graphics and Signal Processing},
  11:21--29.

\bibitem[Ashmawi et~al., 2019]{ASHMAWI2019}
Ashmawi, S., Alharbi, M., Almaghrabi, A., and Alhothali, A. (2019).
\newblock Fitme: Body measurement estimations using machine learning method.
\newblock {\em Procedia Computer Science}, 163:209--217.
\newblock 16th Learning and Technology Conference 2019Artificial Intelligence
  and Machine Learning: Embedding the Intelligence.

\bibitem[Dao et~al., 2014]{Dao2014}
Dao, N.-L., Deng, T., and Cai, J. (2014).
\newblock Fast and automatic human body circular measurement based on a single
  kinect.

\bibitem[Gonzalez-Tejeda and Mayer, 2019]{Tejeda2019}
Gonzalez-Tejeda, Y. and Mayer, H. (2019).
\newblock Calvis: Chest, waist and pelvis circumference from 3d human body
  meshes as ground truth for deep learning.
\newblock In {\em Proceedings of the VIII International Workshop on
  Representation, analysis and recognition of shape and motion FroM Imaging
  data (RFMI 2019)}. ACM.

\bibitem[Guilló et~al., 2020]{Guillo2020}
Guilló, A., Azorin-Lopez, J., Saval-Calvo, M., Castillo-Zaragoza, J.,
  Garcia-D'Urso, N., and Fisher, R. (2020).
\newblock Rgb-d-based framework to acquire, visualize and measure the human
  body for dietetic treatments.
\newblock {\em Sensors}, 20:3690.

\bibitem[Jensen et~al., 2021]{Jensen2021}
Jensen, J., Hannemose, M., B{\ae}rentzen, J., Wilm, J., Frisvad, J., and Dahl,
  A. (2021).
\newblock Surface reconstruction from structured light images using
  differentiable rendering.
\newblock {\em Sensors}, 21(4):1--16.

\bibitem[Li et~al., 2013]{Li2013}
Li, Z., Jia, W., Mao, Z.-H., Li, J., Chen, H.-C., Zuo, W., Wang, K., and Sun,
  M. (2013).
\newblock Anthropometric body measurements based on multi-view stereo image
  reconstruction.
\newblock {\em Conference proceedings : ... Annual International Conference of
  the IEEE Engineering in Medicine and Biology Society. IEEE Engineering in
  Medicine and Biology Society. Conference}, 2013:366--369.

\bibitem[Liu et~al., 2017]{Liu2017}
Liu, K., Wang, J., Kamalha, E., Li, V., and Zeng, X. (2017).
\newblock Construction of a prediction model for body dimensions used in
  garment pattern making based on anthropometric data learning.
\newblock {\em The Journal of The Textile Institute}, 108:1--8.

\bibitem[Loper et~al., 2015]{SMPL2015}
Loper, M., Mahmood, N., Romero, J., Pons-Moll, G., and Black, M.~J. (2015).
\newblock {SMPL}: A skinned multi-person linear model.
\newblock {\em ACM Trans. Graphics (Proc. SIGGRAPH Asia)}, 34(6):248:1--248:16.

\bibitem[Qi et~al., 2016]{PointNet2016}
Qi, C.~R., Su, H., Mo, K., and Guibas, L.~J. (2016).
\newblock Pointnet: Deep learning on point sets for 3d classification and
  segmentation.
\newblock {\em CoRR}, abs/1612.00593.

\bibitem[Rakotosaona et~al., 2019]{Rakotosaona2019}
Rakotosaona, M., Barbera, V.~L., Guerrero, P., Mitra, N.~J., and Ovsjanikov, M.
  (2019).
\newblock {POINTCLEANNET:} learning to denoise and remove outliers from dense
  point clouds.
\newblock {\em CoRR}, abs/1901.01060.

\bibitem[Robinette et~al., 2002]{Robinette2002}
Robinette, K., Blackwell, S., Daanen, H., Boehmer, M., and Fleming, S. (2002).
\newblock Civilian american and european surface anthropometry resource
  (caesar), final report. volume 1. summary.
\newblock page~74.

\bibitem[Rosman et~al., 2012]{Rosman2012}
Rosman, G., Dubrovina, A., and Kimmel, R. (2012).
\newblock Sparse modeling of shape from structured light.
\newblock pages 456--463.

\bibitem[Shigeki et~al., 2018]{Shigeki2018}
Shigeki, Y., Okura, F., Mitsugami, I., and Yagi, Y. (2018).
\newblock Estimating 3d human shape under clothing from a single rgb image.
\newblock {\em IPSJ Transactions on Computer Vision and Applications}, 10:16.

\bibitem[Song et~al., 2017]{Song2017}
Song, D., Tong, R., Chang, J., Wang, T., Du, J., Tang, M., and Zhang, J.
  (2017).
\newblock Clothes size prediction from dressed-human silhouettes.
\newblock pages 86--98.

\bibitem[Tejeda and Mayer, 2021]{Tejeda2021}
Tejeda, Y.~G. and Mayer, H.~A. (2021).
\newblock A neural anthropometer learning from body dimensions computed on
  human 3d meshes.

\bibitem[Tsoli et~al., 2014]{Tsoli2014}
Tsoli, A., Loper, M., and Black, M. (2014).
\newblock Model-based anthropometry: Predicting measurements from 3d human
  scans in multiple poses.

\bibitem[Wang et~al., 2019]{Wang2019}
Wang, Z., Wang, J., Xing, Y., Yang, Y., and LIU, K. (2019).
\newblock Estimating human body dimensions using rbf artificial neural networks
  technology and its application in activewear pattern making.
\newblock {\em Applied Sciences}, 9:1140.

\bibitem[Xiaohui et~al., 2018]{XIAOHUI2018}
Xiaohui, T., Xiaoyu, P., Liwen, L., and Qing, X. (2018).
\newblock Automatic human body feature extraction and personal size
  measurement.
\newblock {\em Journal of Visual Languages \& Computing}, 47:9--18.

\bibitem[Yan and K{\"a}m{\"a}r{\"a}inen, 2021]{Yan2021}
Yan, S. and K{\"a}m{\"a}r{\"a}inen, J.-K. (2021).
\newblock Learning anthropometry from rendered humans.
\newblock {\em ArXiv}, abs/2101.02515.

\bibitem[Yan et~al., 2020]{Yan2020}
Yan, S., Wirta, J., and Kämäräinen, J.-K. (2020).
\newblock Anthropometric clothing measurements from 3d body scans.
\newblock {\em Machine Vision and Applications}, 31.

\end{thebibliography}

%\section*{\uppercase{Appendix}}

\end{document}